\documentclass[10pt,twocolumn,letterpaper]{article}

\usepackage{cvpr}              

\definecolor{cvprblue}{rgb}{0.21,0.49,0.74}
\usepackage[pagebackref,breaklinks,colorlinks,allcolors=cvprblue]{hyperref}

\def\paperID{3039} 
\def\confName{CVPR}
\def\confYear{2025}

\title{UniGoal: Towards Universal Zero-shot Goal-oriented Navigation}

\author{
  Hang Yin$^{1}$\thanks{~Equal contribution. $^\dag$ Project lead. $^\ddagger$ Corresponding author.}, Xiuwei Xu$^{1*\dagger}$, Linqing Zhao$^1$, Ziwei Wang$^2$, Jie Zhou$^1$, Jiwen Lu$^{1\ddagger}$ \\
  $^1$Tsinghua University \\
  $^2$Nanyang Technological University \\
  {\tt\small \{yinh23, xxw21, zhaolinqing\}@mails.tsinghua.edu.cn;} \\
 \tt\small ziwei.wang@ntu.edu.sg; {\tt\small \{jzhou, lujiwen\}@tsinghua.edu.cn} \\
}

\usepackage[utf8]{inputenc} 
\usepackage[T1]{fontenc}    
\usepackage{hyperref}       
\usepackage{url}            
\usepackage{booktabs}       
\usepackage{amsfonts}       
\usepackage{nicefrac}       
\usepackage{microtype}      
\usepackage{xcolor}         
\usepackage{amsmath} 
\usepackage{arydshln}
\usepackage{wrapfig}
\usepackage{color}
\usepackage{multirow}
\usepackage{multicol}
\usepackage{colortbl}
\usepackage{algorithm}
\definecolor{mygray}{gray}{.85}
\definecolor{myhighlight}{RGB}{193,210,240}
\definecolor{newnodepurple}{RGB}{184,170,237}
\definecolor{longedgered}{RGB}{192,0,0}
\definecolor{shortedgegreen}{RGB}{0,176,80}
\definecolor{verylightgray}{RGB}{240,240,240} 
\definecolor{darkred}{RGB}{139, 0, 0}   
\definecolor{darkgreen}{RGB}{0, 100, 0}   
\definecolor{darkblue}{RGB}{0, 0, 139} 
\definecolor{darkredtransparent}{RGB}{190, 120, 120}   
\usepackage{adjustbox}
\usepackage{wrapfig}
\usepackage{listings}  
\usepackage[framemethod=TikZ]{mdframed}  
\usepackage{algorithm}
\usepackage{algorithmic}
\usepackage[most]{tcolorbox}  
\usepackage{graphicx}
\usepackage{subcaption}
\usepackage{array}
\newcolumntype{M}[1]{>{\centering\arraybackslash}m{#1}}

\definecolor{shapecolor}{rgb}{0.0,0.5,0.0}
\definecolor{mygray}{gray}{.85}

\usepackage{listings}
\usepackage{color}
\newmdenv[  
  backgroundcolor=verylightgray,  
  hidealllines=true,  
  innerleftmargin=8pt,  
  innerrightmargin=8pt,  
  innertopmargin=2pt,  
  innerbottommargin=4pt  
]{graybox}

\begin{document}
\maketitle
\begin{abstract}
    In this paper, we propose a general framework for universal zero-shot goal-oriented navigation. Existing zero-shot methods build inference framework upon large language models (LLM) for specific tasks, which differs a lot in overall pipeline and fails to generalize across different types of goal. 
Towards the aim of universal zero-shot navigation, we propose a uniform graph representation to unify different goals, including object category, instance image and text description. We also convert the observation of agent into an online maintained scene graph. With this consistent scene and goal representation, we preserve most structural information compared with pure text and are able to leverage LLM for explicit graph-based reasoning.
Specifically, we conduct graph matching between the scene graph and goal graph at each time instant and propose different strategies to generate long-term goal of exploration according to different matching states. The agent first iteratively searches subgraph of goal when zero-matched. With partial matching, the agent then utilizes coordinate projection and anchor pair alignment to infer the goal location. Finally scene graph correction and goal verification are applied for perfect matching. We also present a blacklist mechanism to enable robust switch between stages.
Extensive experiments on several benchmarks show that our UniGoal achieves state-of-the-art zero-shot performance on three studied navigation tasks with a single model, even outperforming task-specific zero-shot methods and supervised universal methods. \href{https://bagh2178.github.io/UniGoal/}{bagh2178.github.io/UniGoal/}
\end{abstract}   

\section{Introduction}
    Goal-oriented navigation is a fundamental problem in various robotic tasks, which requires the agent to navigate to a specified goal in an unknown environment. Depending on the goal type, there are many popular sub-tasks of goal-oriented navigation, among which we focus on three representative types: object category, instance image and text description. These sub-tasks are also known as Object-goal Navigation (ON)~\cite{chaplot2020object,zhou2023esc,yin2024sg}, Instance-image-goal Navigation (IIN)~\cite{krantz2022instance,krantz2023navigating} and Text-goal Navigation (TN)~\cite{sun2024prioritized}.
With the development of deep learning, reinforcement learning (RL) and vision/language foundation models, we have witnessed great achievement of performance on each individual sub-task. 
However, in actual application scenarios, the high flexibility of human instructions requires high versatility of agent. Therefore, a universal method that can handle all sub-tasks in a single model is of great desire.

\begin{figure}
    \centering
    \includegraphics[width=\linewidth]{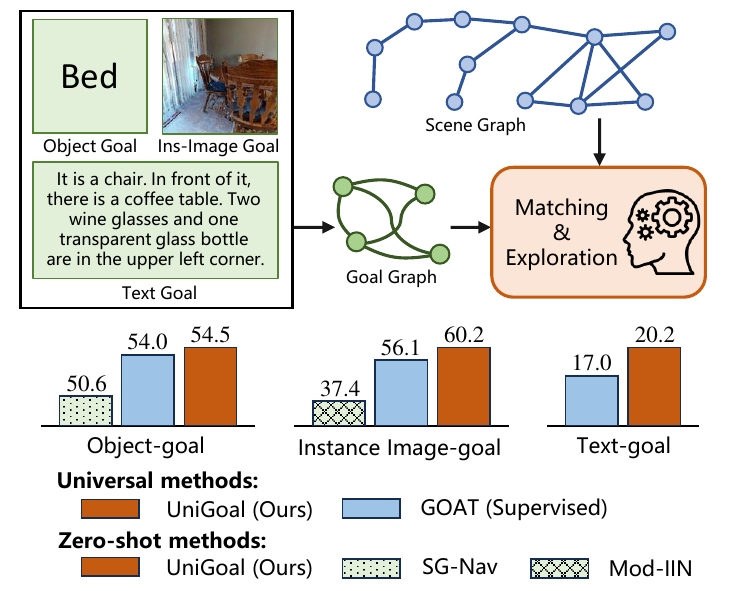}
    \caption{State-of-the-art zero-shot goal-oriented navigation methods are typically specialized for each goal type. Although recent work presents universal goal-oriented navigation method, it requires to train policy networks on large-scale data and lacks zero-shot generalization ability. We propose \textbf{UniGoal}, which enables zero-shot inference on three studied navigation tasks with a unified framework and achieves leading performance on multiple benchmarks.}
    \label{fig:teaser}
\end{figure}

Towards the aim of universal goal-oriented navigation, a natural solution is to learn a uniform representation of different kinds of goals. GOAT~\cite{chang2023goat} trains a universal global policy on the three goal-oriented sub-tasks, which learns a shared goal embedding with RL. To 
reduce the requirement of training resources, PSL~\cite{sun2024prioritized} utilizes CLIP~\cite{radford2021learning} embedding to uniformly represent category, image and text description. But it still requires time-consuming RL training for the policy networks. Moreover, these training-based methods tend to overfit on the simulation environment and thus show weak generalization ability when applied to real world. 
To solve above limitations, zero-shot navigation methods~\cite{zhou2023esc,yin2024sg} appear to be an ideal choice, where the agent does not require any training or finetuning when deployed to a certain task.
The mainstream solution of zero-shot methods is to leverage large language models (LLM)~\cite{touvron2023llama,achiam2023gpt} for general reasoning and decision making.
However, although utilizing LLM makes zero-shot navigation feasible, current methods are designed for specific sub-task, which cannot transfer to wider range of goal types. The recent InstructNav~\cite{long2024instructnav} proposes a general framework to solve several language-related navigation tasks with chain-of-thought, but it is still unable to handle vision-related navigation like IIN. Therefore, a uniform inference framework for universal zero-shot goal-oriented navigation is highly demanded.

In this paper, we propose UniGoal to solve the above problems. Different from previous works which represent scene and goal in text format and design task-specific workflow for LLM, we propose a uniform graph representation for both 3D scene and goal and formulate a general LLM-based scene exploration framework.
With our graph-based representation, the 3D scene, object category, instance image and text description can be uniformly represented with minimal structural information loss compared to text description. The consistent graph format between scene and goal also enables accurate explicit reasoning including similarity computation, graph matching and graph alignment.
Specifically, we construct an online 3D scene graph along with the moving of agent. At each time instant, we first conduct graph matching between the scene graph and goal graph. Then we propose a multi-stage scene exploration policy, which adopts different strategies to generate long-term goal of exploration according to different matching states. With exploration of unknown regions, the matching score will increase and the policy will progress between three stages: iterative subgraph searching for zero matching, coordinate projection and anchor pair alignment for partial matching, and scene graph correction and goal verification for perfect matching. 
To enable robust switch between stages, we also present a blacklist mechanism to freeze unmatched parts of graphs and encourage exploration to new regions.
Experimental results on several benchmarks of MatterPort3D~\cite{Matterport3D}, HM3D~\cite{ramakrishnan2021habitatmatterport} and RoboTHOR~\cite{RoboTHOR} show that UniGoal achieves superior performance on all three tasks with a single model, even outperforming zero-shot methods that designed for specific task and universal methods that requires training or finetuning.

\section{Related Work}
    \textbf{Zero-shot Navigation.} Conventional supervised navigation methods~\cite{chaplot2020object, ramakrishnan2022poni, wijmans2019dd, Kwon_2023_CVPR, du2021curious} requires large-scale training in simulation environments, which limits the generalization ability. According to the goal type, zero-shot navigation can be mainly divided into ON~\cite{yu2023l3mvn, cai2024bridging}, IIN~\cite{krantz2023navigating} and TN~\cite{long2024instructnav}. 
Based on open-vocabulary CLIP~\cite{radford2021learning}, CoW~\cite{gadre2023cows} constructs zero-shot ON baseline using frontier-based exploration (FBE). ESC~\cite{zhou2023esc}, OpenFMNav~\cite{chen2023open} and VLFM~\cite{yokoyama2024vlfm} further extract common sense about correlations between objects using LLM for goal location reasoning. 
For zero-shot IIN, Mod-IIN~\cite{krantz2023navigating} simply utilizes FBE for exploration and key point matching for goal identification.
For TN, currently there are only supervised methods~\cite{chang2023goat,sun2024prioritized} and zero-shot ones are still missing. The recent InstructNav~\cite{long2024instructnav} proposes a universal zero-shot framework for language-related navigation tasks. It can be applied to ON, demand-driven navigation (DDN) and vision-language-navigation (VLN)~\cite{chen2024mapgpt, zhou2024navgpt, long2024discuss, li2024tina, zhan2024mc}. But it is still unable to handle visual goal as in IIN.
Different from these approaches, our UniGoal proposes a unified graph representation for goal and scene, which can elegantly handle both language and vision-related goal-oriented navigation tasks.

\noindent\textbf{Graph-based Scene Exploration.} In order to better understand and explore the scene, there are a variety of scene representation methods. Among them, graph-based representation~\cite{gu2023conceptgraphs, armeni20193dscenegraphstructure, kim20193, rosinol20203ddynamicscenegraphs, wald2020learning3dsemanticscene} is one of the most popular and promising representations based on explicit graph structure, which shows great potential to be combined with LLM and VLM for high-level reasoning.
SayPlan~\cite{rana2023sayplan} leverages LLM to perform task planning by searching on a 3D scene graph. 
OVSG~\cite{pmlr-v229-chang23b} constructs an open-vocabulary scene graph for context-aware descriptions and performs graph matching to ground the queried entity.
There are also many works using graph representation for navigation tasks~\cite{wu2024voronav,rajvanshi2024saynav,liu2023bird,yin2024sg}.
Among them the most related work to ours is SG-Nav~\cite{yin2024sg}, which constructs an online hierarchical scene graph and proposes chain-of-thought prompting for LLM to reason the goal location based on graph sturcture.
However, SG-Nav is specialized for ON, thus cannot fully exploit the rich information contained in other types of goal. While our UniGoal further represents the goal in a graph and perform graph matching between scene and goal to guide a multi-stage scene exploration policy, which make full use of the correlation between scene and goal for LLM reasoning.

\section{Approach}
    \begin{figure*}
    \centering
    \includegraphics[width=\linewidth]{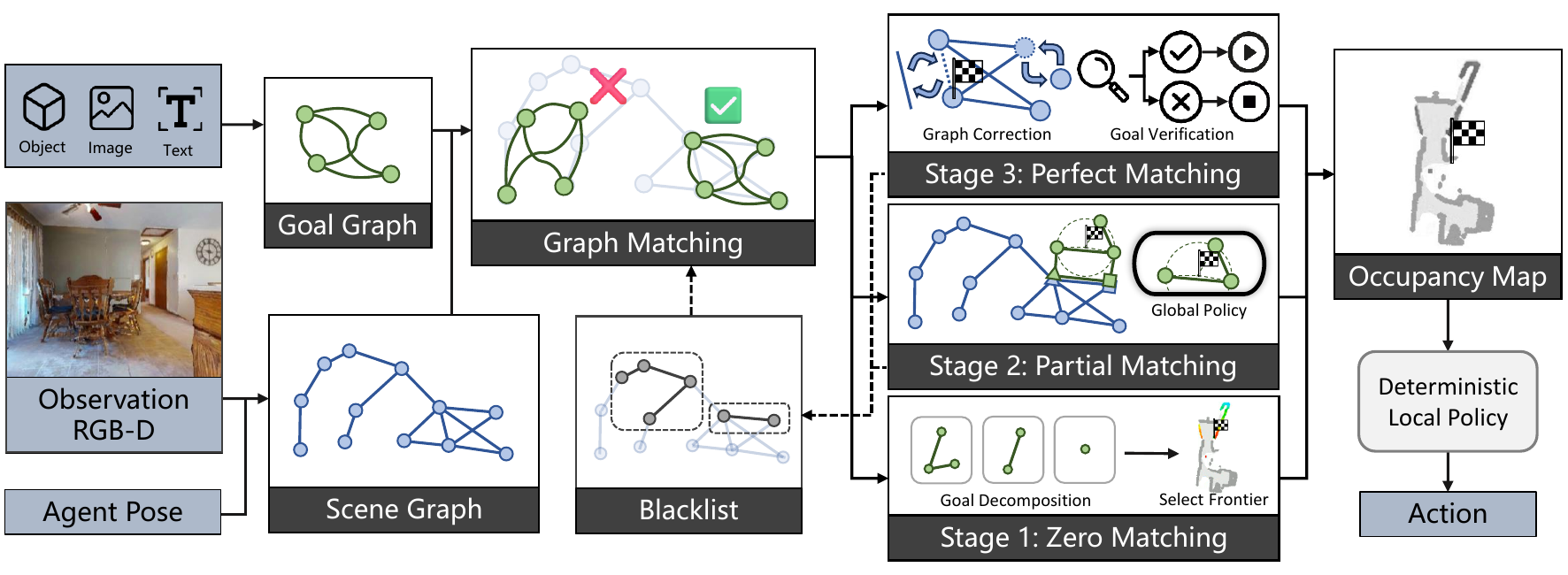}
    \caption{Framework of UniGoal. We convert different types of goals into a uniform graph representation and maintain an online scene graph. At each step, we perform graph matching between the scene graph and goal graph, where the matching score will be utilized to guide a multi-stage scene exploration policy. For different degree of matching, our exploration policy leverages LLM to exploit the graphs with different aims: first expand observed area, then infer goal location based on the overlap of graphs, and finally verify the goal. We also propose a blacklist that records unsuccessful matching to avoid repeated exploration.}
    \label{fig:framework}
\end{figure*}

In this section, we first provide the definition of universal zero-shot goal-oriented navigation and introduce the framework of UniGoal. Then we construct graphs for scene and goal and conduct graph matching for goal identification. Next we design a multi-stage scene exploration policy by prompting LLM with graphs. Finally we propose a blacklist mechanism to robustly avoid repeated exploration.

\subsection{Goal-oriented Navigation}
In goal-oriented navigation, a mobile agent is tasked with navigating to a specified goal $g$ in an unknown environment, where $g$ can be an object category (i.e.\ Object-goal Navigation~\cite{chaplot2020object}, ON), an image containing object that can be found in the scene (i.e.\ Instance-Image-goal Navigation~\cite{krantz2022instance}, IIN), or a description about a certain object (i.e.\ Text-goal Navigation~\cite{sun2024prioritized}, TN). 
For IIN and TN, there is a central object $o$ as well as other relevant objects in $g$. While for ON, we have $o=g$.
The agent receives posed RGB-D video stream and is required to execute an action $a\in \mathcal{A}$ at each time it receiving a new RGB-D observation. $\mathcal{A}$ is the set of actions, which consists of \texttt{move\_forward}, \texttt{turn\_left}, \texttt{turn\_right} and \texttt{stop}. The task is successfully done if the agent stops within $r$ meters of $o$ in less than $T$ steps. More details about the three sub-tasks can be found in supplementary material.

\textbf{Task Specification.} We aim to study the problem of universal zero-shot goal-oriented navigation, which has two characteristics: (1) Universal. We should design a general method, which requires no modification when switching between the three sub-tasks. (2) Zero-shot. All three kinds of goal can be specified by free-form language or image. Our navigation method does not require any training or finetuning, which is of great generalization ability.

\textbf{Overview.} Universal zero-shot goal-oriented navigation requires the agent to complete different sub-tasks with a single framework in training-free manner. Since this task requires extremely strong generalization ability, we utilize large language model (LLM)~\cite{touvron2023llama,achiam2023gpt} for zero-shot decision making by exploiting its rich knowledge and strong reasoning ability. 
To make LLM aware of the visual observations as well as unifying different kinds of goals, we propose to represent the scene and goal in graphs, i.e., \emph{scene graph} and \emph{goal graph}. In this way, different goals are represented uniformly and the representations of scene and goal are consistent.
Based on this representation, we prompt LLM with scene graph and goal graph for scene understanding, graph matching and decision making for exploration. The overall pipeline is illustrated in Figure \ref{fig:framework}.

\subsection{Graph Construction and Matching}
In this subsection, we first describe how to construct a uniform graph representation for scene and goal. Then we propose a graph matching method to determine whether a goal or its relevant objects are observed, which further guides the selection of scene exploration policies.

\textbf{Graph Construction.} We define graph $\mathcal{G}=(\mathcal{V}, \mathcal{E})$ as a set of nodes $\mathcal{V}$ connected with edges $\mathcal{E}$. Each node represents an object. Each edge represents the relationship between objects, which only exists between spatially or semantically related object pairs. The content of nodes and edges is described in text format.
Since the agent is initialized in unknown environment and continuously explores the scene, we follow SG-Nav~\cite{yin2024sg} to construct the scene graph $\mathcal{G}_t$ incrementally by expanding it every time the agent receives a new RGB-D observation. 
For goal graph $\mathcal{G}_g$, we adopt different methods to process three kinds of goals $g$ into a graph, which we detail in supplementary material. 


\textbf{Graph Matching.} With scene graph $\mathcal{G}_t$ and goal graph $\mathcal{G}_g$, we can match these two graphs to determine whether the goal is observed. If no elements in $\mathcal{G}_g$ are observed, the agent needs to infer relationship between objects from $\mathcal{G}_t$ to plan a path that is most likely to find the goal. If $\mathcal{G}_g$ is partially observed in $\mathcal{G}_t$, the agent can use the overlapped part of $\mathcal{G}_g$ and $\mathcal{G}_t$ to reason out where the rest of $\mathcal{G}_g$ is. If $\mathcal{G}_g$ is perfectly observed in $\mathcal{G}_t$, the agent can move to the goal and make further verification. Therefore, a goal scoring method is crucial for the follow-up scene exploration.

We propose to apply graph matching to achieve this. Given $\mathcal{G}_t$ and $\mathcal{G}_g$, we design three matching metrics, i.e., node matching, edge matching and topology matching, to score how well the goal is observed. Formally, for nodes and edges, we extract their embeddings and then compute pair-wise similarity with bipartite matching to determine matched pairs of nodes and edges:
\begin{align}
    \mathcal{M}_N&=\mathcal{B}({\rm thr}({\rm Embed}(\mathcal{V}_t)\cdot {\rm Embed}(\mathcal{V}_g)^T)) \\
    \mathcal{M}_E&=\mathcal{B}({\rm thr}({\rm Embed}(\mathcal{E}_t)\cdot {\rm Embed}(\mathcal{E}_g)^T))
\end{align}
where ${\rm Embed}(\cdot)\in \mathbb{R}^{K\times C}$, $K$ is the number of nodes or edges and $C$ is the channel dimension, which is detailed in supplementary material.
${\rm thr}(\cdot)$ is an element-wise threshold function applied on the similarity matrix which sets values smaller than $\tau$ to -1 to disable matching of the corresponding pairs. $\mathcal{B}(\cdot)$ is bipartite matching, which outputs a list of all matched node or edge pairs, namely $\mathcal{M}_N$ or $\mathcal{M}_E$.
We also average the similarity matrix of nodes and edges to acquire the similarity scores $S_N$ and $S_E$.

Based on $\mathcal{M}_N$ and $\mathcal{M}_E$, we further compute topological similarity between $\mathcal{G}_t$ and $\mathcal{G}_g$, which is defined as the graph editing similarity between them:
\begin{equation}
    S_T=1-\mathcal{D}(\mathcal{S}(\mathcal{F}(\mathcal{G}_t, \mathcal{M}_N, \mathcal{M}_E)), \mathcal{S}(\mathcal{G}_g))
\end{equation}
where $\mathcal{F}(\mathcal{G}_t, \mathcal{M}_N, \mathcal{M}_E)$ means the minimal subgraph of $\mathcal{G}_t$ with nodes in $\mathcal{M}_N$ and edges in $\mathcal{M}_E$. $\mathcal{S}(\cdot)$ means the topological structure of a graph regardless of the content of nodes and edges. $\mathcal{D}(\cdot)$ is the normalized editing distance~\cite{sanfeliu1983distance} between two graphs. The final matching score is defined as $S=(S_N+S_E+S_T)/3$.

\subsection{Multi-stage Scene Exploration}
As described above, different matching scores will lead to different scene exploration policies. From zero matching to perfect matching, we design three stages to progressively explore the scene and generate long-term exploration goal. This long-term goal will be processed by a deterministic local policy~\cite{sethian1999fast} to obtain actions. Below we detail our exploration policy stage by stage.

\textbf{Stage 1: Zero Matching.}
If the matching score $S$ is smaller than $\sigma_1$, we regard this stage as zero matching. Since there is almost no element of $\mathcal{G}_g$ observed in $\mathcal{G}_t$, the aim of agent at this stage is to expand its explored region and find elements in $\mathcal{G}_g$.
Note this problem is similar to ON: before observing the goal, the agent needs to explore unknown regions without any matching between goal and scene. We can simply resort to scene graph-based ON method~\cite{yin2024sg} at this stage, which navigates to frontiers with semantic relationships between the scene graph and goal as guidance.

However, different from ON where the goal is always an object node, in our universal goal-oriented navigation the goal may be a complicated graph. A graph may consist of several less related subgraph parts. For example, in a graph (table, chair, window, curtain), [table, chair] and [window, curtain] are two subgraphs which have strong internal correlation but weak interrelation.
We empirically observe that locating a collection of multiple unrelated subgraphs in a scene at the same time is much more difficult than locating a single subgraph once at a time.
Therefore, we decompose $\mathcal{G}_g$ into multiple internally correlated subgraphs with the guidance of LLM. For each subgraph of $\mathcal{G}_g$, we convert it to a text description, which is regarded as an object goal to call \cite{yin2024sg} for frontier selection.
Finally, we select one frontier from the proposed ones by averaging the frontier scores and distances to the agent.
Details about LLM-guided decomposition and score computation can be found in supplementary material. 
With this strategy, we not only utilize the information of the entire $\mathcal{G}_g$, but also eliminate ambiguity during frontier selection caused by unrelated subgraphs.

\textbf{Stage 2: Partial Matching.}
With the exploration of agent, the elements of $\mathcal{G}_g$ will gradually be observed in $\mathcal{G}_t$ and thus $S$ continues to increase. When $S$ exceeds $\sigma_1$ (but still smaller than $\sigma_2$) and there is at least one anchor pair, we switch to partial matching stage. Here anchor pair means a pair of exactly matched nodes, i.e., two unconnected matched nodes in $\mathcal{M}_N$ or one matched edge in $\mathcal{M}_M$.

Note that we store the world coordinates of nodes in $\mathcal{G}_t$. If we also know the relative coordinates of nodes in $\mathcal{G}_g$, we can map the coordinates of $\mathcal{G}_t$ and $\mathcal{G}_g$ to bird's-eye view (BEV) and align the anchor pair of $\mathcal{G}_g$ to the one of $\mathcal{G}_t$. In this way, after alignment we can directly infer where the rest of $\mathcal{G}_g$ is, which provides the agent with clear exploration goal for each anchor pair. 
Luckily, although we do not have any coordinate information about the goal, at least we are aware of the relative spatial relationship between nodes in $\mathcal{G}_g$, such as a chair \emph{on the left of} a table, a keyboard \emph{in front of} a monitor.
Inspired by this, we propose a coordinate projection strategy that preserves spatial relationships.

\begin{figure}
    \centering
    \includegraphics[width=\linewidth]{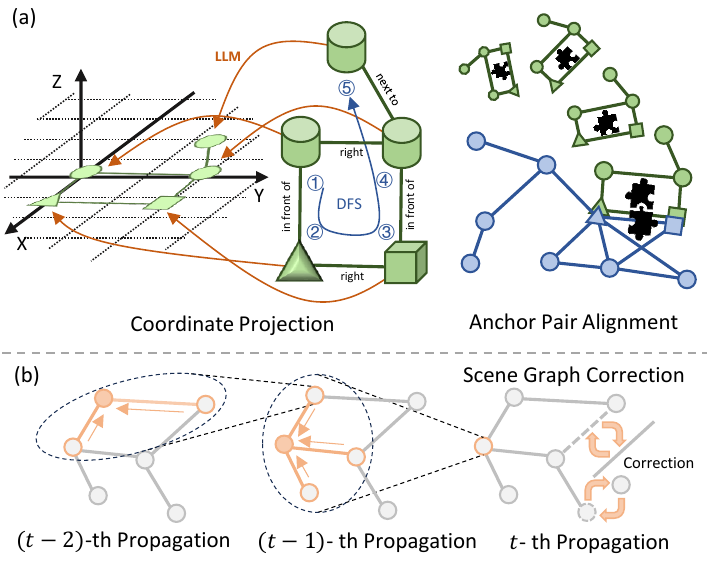}
    \caption{Illustration of approach. (a) Stage 2: coordinate projection and anchor pair alignment. (b) Stage 3: scene graph correction.}
    \label{fig:approach}
\end{figure}

Given $\mathcal{G}_g$ without any coordinate information, we first project the central object node $o$ to $(0,0)$ as an initialization. To infer the projected coordinates of other nodes, we need to utilize the spatial relationship between coordinate-known nodes and coordinate-unknown nodes. Since at beginning we only have one coordinate-known node $o$, we start from it and traverse the goal graph with Depth First Search (DFS) to gradually infer the projected BEV coordinates of the whole graph. During the traversal, we focus on the edge connecting last node (coordinate-known) and current node (coordinate-unknown), which stores the spatial relationship. For each edge, we prompt LLM with:
\textcolor{brown}{\textit{[Current Node] is [Relationship] [Last Node], coordinate of [Last Node] is $(x,y)$, the X-axis and Y-axis are positive towards yourself and the right respectively. What is the coordinate of [Current Node]?}}
to project the current coordinate-unknown node to BEV.

With the projected $\mathcal{G}_t$ and $\mathcal{G}_g$, we conduct alignment for each anchor pair in order. Assume that the anchors are $\mathbf{v}_{t}^{1},\mathbf{v}_{t}^{2} \in \mathcal{G}_t$ and $\mathbf{v}_{g}^{1},\mathbf{v}_{g}^{2} \in \mathcal{G}_g$. $\mathbf{P} \in \mathbb{R}^{3 \times 3}$ is the 2D coordinate transfer matrix, consisting of scale $\mathbf{S}$, rotation $\mathbf{R}$ and translation $\mathbf{T}$, which is formulated as:
\begin{equation}
    \mathbf{P}=\mathbf{S} \cdot \mathbf{R} \cdot \mathbf{T}=\left[\begin{array}{ccc}
    s \cos (\theta) & -s \sin (\theta) & t_{x} \\
    s \sin (\theta) & s \cos (\theta) & t_{y} \\
    0 & 0 & 1
    \end{array}\right]
\end{equation}
Based on the anchor pair relationship, we establish the equation $\mathbf{v}_{t}^{1}=\mathbf{P} \cdot \mathbf{v}_{g}^{1}$ and $\mathbf{v}_{t}^{2}=\mathbf{P} \cdot \mathbf{v}_{g}^{2}$. Then we can solve the parameters $t_{x},t_{y},\theta,s$ in $\mathbf{P}$.
With the coordinate transfer matrix $\mathbf{P}$, we project the rest nodes $\mathbf{v}_{g} \subset \mathcal{V}_{g}$ into the coordinate of $\mathcal{G}_t$ as $\mathbf{v}_{t}=\mathbf{P} \cdot \mathbf{v}_{g}$. Finally, the exploration goal for this anchor pair can be set as the center $\mathbf{c}^{*}$ of the smallest circumcircle of the projected nodes:
\begin{equation}
    \mathbf{c}^{*} = \text{argmin}_{\mathbf{c}} \left\{ r \mid \forall \mathbf{v}\in \mathbf{v}_t, \|\mathbf{v} - \mathbf{c}\| \leq r \right\}
\end{equation}
which ensures that the distance from the exploration goal to the farthest node in $\mathbf{v}_{t}$ is minimal.

\textbf{Stage 3: Perfect Matching.}
When the matching score $S$ exceeds $\sigma_2$ and the central object $o$ of $\mathcal{G}_g$ is matched ($o\in \mathcal{M}_N$), we switch the exploration policy to perfect matching stage. Since $o$ is matched to an observed node in $\mathcal{G}_t$, we can simply have the agent move to this object without further exploration. However, there may be perceptual errors during scene graph construction. To ensure the matched $o$ is correct, we propose a graph correction and goal verification pipeline to refine unreasonable structure in $\mathcal{G}_t$ as well as judge the confidence of goal in the process of approaching $o$.

We include nodes and edges of $\mathcal{G}_t$ within distance $d$ of $o$ into the correction scope, which is defined as a subgraph $\mathcal{G}_o=(\mathcal{V}_{o},\mathcal{E}_{o})$ with $n$ nodes and $m$ edges.
As the agent approaches $o$, it continuously receives new RGB-D observation $\mathcal{I}^{(t)}$ and utilizes both visual observation and graph relationship to correct $\mathcal{G}_o$. Similar to graph convolution, we propagate information from neighbor nodes and edges and utilize LLM for information aggregation and updating:
\begin{align}
    \mathcal{V}_{o}^{(t+1)}&=\text{LLM}(\mathbf{A} \cdot \mathcal{V}^{(t)}_{o},\mathbf{M} \cdot \mathcal{E}_{o}^{(t)},\text{VLM}(\mathcal{I}^{(t)})) \\
    \mathcal{E}_{o}^{(t+1)}&=\text{LLM}(\mathbf{M}^{T} \cdot \mathcal{V}^{(t)}_{o},\mathbf{A}' \cdot \mathcal{E}_{o}^{(t)},\text{VLM}(\mathcal{I}^{(t)}))
\end{align}
where $\mathbf{A} \in \mathbb{R}^{n \times n}$, $\mathbf{A}' \in \mathbb{R}^{m \times m}$ and $\mathbf{M} \in \mathbb{R}^{n \times m}$ are the adjacency matrix, edge adjacency matrix and incidence matrix of $\mathcal{G}_o$.
$\mathcal{V}_{o}^{(t)}$ and $\mathcal{E}_{o}^{(t)}$ are the description of nodes and edges after $t$-th propagation.
After several iterations, we prompt LLM to update unreasonable nodes and edges based on the aggregated $\mathcal{V}_{o}^{(t)}$ and $\mathcal{E}_{o}^{(t)}$. We detail the prompts in supplementary material.

To verify the goal, we consider several confidence items since entering stage 3 as follows:
\begin{equation}
    C_t=N_t+M_t+S_t-\lambda D_t
\end{equation}
where $N_t, M_t, S_t$ are the proportions of corrected nodes and edges, matched keypoints using LightGlue~\cite{lindenberger2023lightglue} (for IIN only) and graph matching score at time $t$. $D_t$ is the path length since stage 3. If $C_t$ exceeds a threshold $C_{thr}$ within $t$ steps, $o$ will be verified. Otherwise $o$ will be excluded.

\subsection{Robust Blacklist Mechanism}
Since our graph matching method always outputs matched result which maximizes the similarity score, if one matching between $\mathcal{G}_t$ and $\mathcal{G}_g$ finally fails to find the goal, relevant nodes and edges in $\mathcal{G}_t$ should be marked to avoid repeated matching.
To this end, we present a blacklist mechanism to record the unsuccessful matching.

Blacklist is initialized as empty. The nodes and edges in blacklist will not be considered in our graph matching method. 
Two cases will extend the blacklist: (1) all anchor pairs of stage 2 in a single matching fails to enter stage 3. In this case, the nodes and edges in these anchor pairs will be appended to blacklist; (2) the goal verification of stage 3 fails. This will move all matched pairs in $\mathcal{M}_N$ and $\mathcal{M}_E$ to blacklist.
Besides, if any node or edge is refined during scene graph correction of stage 3, these nodes and edges along with their connected ones will be removed from blacklist.

\begin{table*}[t]
    \centering
    \setlength\tabcolsep{7pt}
    \caption{Results of Object-goal navigation, Instance-image-goal navigation and Text-goal navigation on MP3D, HM3D and RoboTHOR. We compare the SR and SPL of state-of-the-art methods in different settings. Universal goal-oriented navigation methods are colored in gray.}
    \resizebox{\textwidth}{!}{
    \begin{tabular}{lccccccccccccc}
        \toprule
        \multirow{3}*{\textbf{Method}} & \multirow{3}*{\textbf{Training-Free}} & \multirow{3}*{\textbf{Universal}} &
        \multicolumn{6}{c}{\textbf{ObjNav}} & \multicolumn{2}{c}{\textbf{InsINav}} & \multicolumn{2}{c}{\textbf{TextNav}}\\
        \cmidrule(lr){4-9} \cmidrule(lr){10-11} \cmidrule(lr){12-13}
        & & & \multicolumn{2}{c}{\textbf{MP3D}} & \multicolumn{2}{c}{\textbf{HM3D}} & \multicolumn{2}{c}{\textbf{RoboTHOR}} & 
        \multicolumn{2}{c}{\textbf{HM3D}} & \multicolumn{2}{c}{\textbf{HM3D}} \\
        \cmidrule(lr){4-5} \cmidrule(lr){6-7} \cmidrule(lr){8-9} \cmidrule(lr){10-11} \cmidrule(lr){12-13}
        & & & SR & SPL & SR & SPL & SR & SPL & SR & SPL & SR & SPL \\
        \midrule
        SemEXP~\cite{chaplot2020object} & $\times$ & $\times$ & 36.0 & 14.4 & -- & -- & -- & -- & -- & -- & -- & -- \\
        ZSON~\cite{majumdar2022zson} & $\times$ & $\times$ & 15.3 & 4.8 & 25.5 & 12.6 & -- & -- & -- & -- & -- & -- \\
        OVRL-v2~\cite{yadav2023ovrl} & $\times$ & $\times$ & -- & -- & 64.7 & 28.1 & -- & -- & -- & -- & -- & -- \\
        Krantz et al.~\cite{krantz2022instance} & $\times$ & $\times$ & -- & -- & -- & -- & -- & -- & 8.3 & 3.5 & -- & -- \\
        OVRL-v2-IIN~\cite{yadav2023ovrl} & $\times$ & $\times$ & -- & -- & -- & -- & -- & -- & 24.8 & 11.8 & -- & -- \\
        IEVE~\cite{lei2024instance} & $\times$ & $\times$ & -- & -- & -- & -- & -- & -- & 70.2 & 25.2 & -- & -- \\
        \rowcolor{mygray}PSL~\cite{sun2024prioritized} & $\times$ & $\checkmark$ & -- & -- & 42.4 & 19.2 & -- & -- & 23.0 & 11.4 & 16.5 & 7.5 \\
        \rowcolor{mygray}GOAT~\cite{chang2023goat} & $\times$ & $\checkmark$ & -- & -- & 50.6 & 24.1 & -- & -- & 37.4 & 16.1 & 17.0 & 8.8 \\
        \midrule
        ESC~\cite{zhou2023esc} & $\checkmark$ & $\times$ & 28.7 & 14.2 & 39.2 & 22.3 & 38.1 & 22.2 & -- & -- & -- & -- \\
        OpenFMNav~\cite{kuang2024openfmnav} & $\checkmark$ & $\times$ & 37.2 & 15.7 & 52.5 & 24.1 & 44.1 & 23.3 & -- & -- & -- & -- \\
        VLFM~\cite{yokoyama2024vlfm} & $\checkmark$ & $\times$ & 36.2 & 15.9 & 52.4 & \textbf{30.3} & 42.3 & 23.0 & -- & -- & -- & -- \\
        SG-Nav~\cite{yin2024sg} & $\checkmark$ & $\times$ & 40.2 & 16.0 & 54.0 & 24.9 & 47.5 & 24.0 & -- & -- & -- & -- \\
        Mod-IIN~\cite{krantz2023navigating} & $\checkmark$ & $\times$ & -- & -- & -- & -- & -- & -- & 56.1 & 23.3 & -- & -- \\
        \rowcolor{mygray}UniGoal & $\checkmark$ & $\checkmark$ & \textbf{41.0} & \textbf{16.4} & \textbf{54.5} & 25.1 & \textbf{48.0} & \textbf{24.2} & \textbf{60.2} & \textbf{23.7} & \textbf{20.2} & \textbf{11.4} \\
        \bottomrule
    \end{tabular}}
    \label{tab:main}
\end{table*}

\section{Experiments}
    In this section, we conduct extensive experiments to validate the effectiveness of UniGoal. We first describe experimental settings. Then we compare UniGoal with state-of-the-art methods and ablate each component in UniGoal. Finally we demonstrate some qualitative results.

\subsection{Benchmarks and Implementation Details}
\textbf{Datasets:} We evaluate UniGoal on object-goal, instance-image-goal and text-goal navigation. 
For ON, we conduct experiments on the widely used Matterport3D (MP3D~\cite{Matterport3D}), Habitat-Matterport 3D (HM3D~\cite{ramakrishnan2021habitatmatterport}) and RoboTHOR~\cite{RoboTHOR} following the setting of SG-Nav~\cite{yin2024sg}.
For IIN and TN, we compare with other methods on HM3D, following Mod-IIN~\cite{krantz2023navigating} and InstanceNav~\cite{sun2024prioritized} respectively.

\noindent\textbf{Evaluation Metrics:} We report \textit{success rate (SR)} and \textit{success rate weighted by path length (SPL)}. SR represents the proportion of successful navigation episodes, while SPL measures how close the path is to the optimal path. 

\noindent\textbf{Compared Methods:} We compare with previous state-of-the-art methods on the studied three tasks. For ON, we compare with the supervised methods SemEXP~\cite{chaplot2020object}, ZSON~\cite{majumdar2022zson}, OVRL-v2~\cite{yadav2023ovrl}, and zero-shot methods ESC~\cite{zhou2023esc}, OpenFMNav~\cite{kuang2024openfmnav}, VLFM~\cite{yokoyama2024vlfm}, SG-Nav~\cite{yin2024sg}. 
For IIN, we compare with the supervised methods Krantz et al.~\cite{krantz2022instance}, OVRL-v2 (implementation from \cite{lei2024instance}), IEVE~\cite{lei2024instance}, and zero-shot methods Mod-IIN~\cite{krantz2023navigating}.
For TN, since currently there is no zero-shot method available, we compare with the supervised methods PSL~\cite{sun2024prioritized} and GOAT~\cite{chang2023goat}. PSL and GOAT are also universal methods, for which we also compare with them on other two tasks.

\noindent\textbf{Implementation Details:} We set up our agent in Habitat Simulator~\cite{savva2019habitat, szot2021habitat}. We deploy LLaMA-2-7B~\cite{touvron2023llama} as the LLM and LLaVA-v1.6-Mistral-7B~\cite{liu2023llava} as the VLM throughout the text. We adopt CLIP~\cite{radford2021learning} text encoder to extract the embedding of nodes and edges during graph matching.
Hyperparameters are detailed in appendix.

\begin{figure*}[t]
    \centering
    \includegraphics[width=\linewidth]{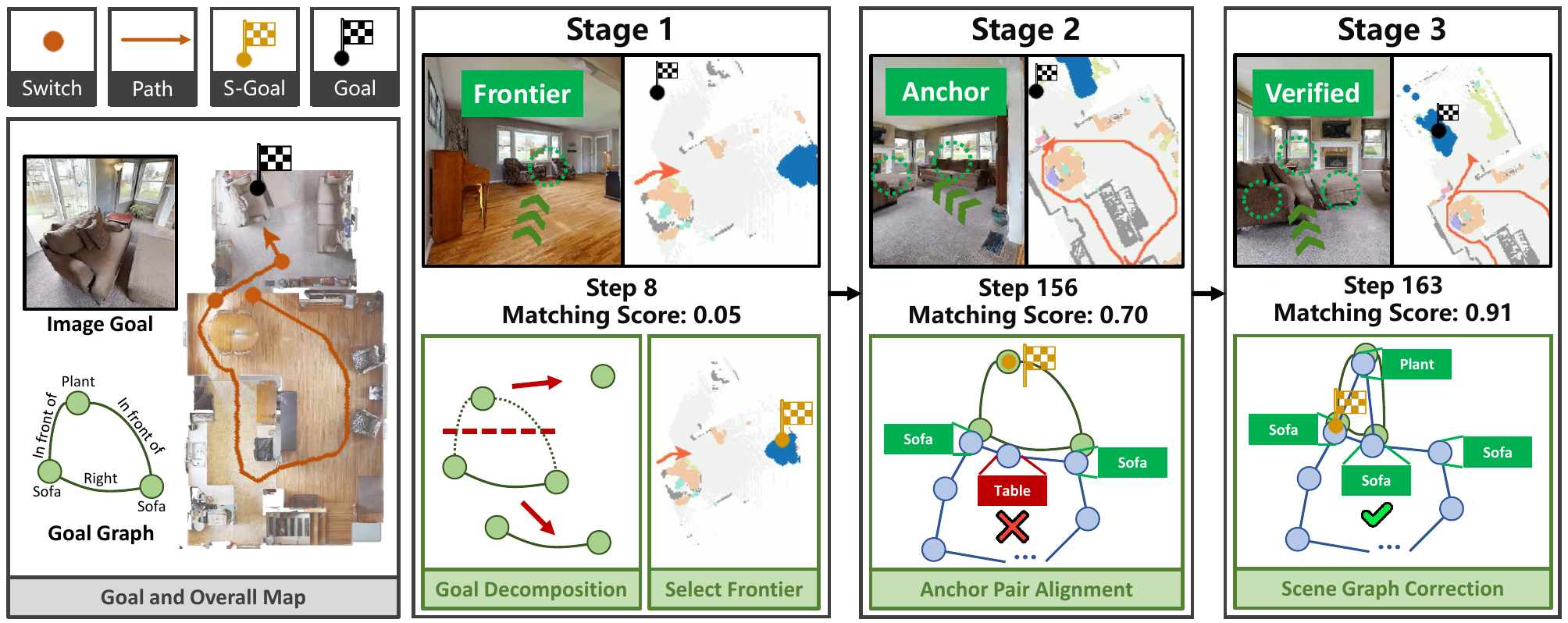}
    \caption{Demonstration of the decision process of UniGoal. Here `Switch' means the point when stage is changing. `S-Goal' means the long-term goal predicted in each stage.}
    \label{fig:visualization_approach}
\end{figure*}

\subsection{Comparison with State-of-the-art}
We compare UniGoal with the state-of-the-art goal-oriented navigation methods of different setting, including supervised, zero-shot and universal methods on three studied tasks in Table \ref{tab:main}. On zero-shot ON and IIN, UniGoal surpasses state-of-the-art methods SG-Nav and Mod-IIN by 0.8\% and 4.1\% respectively. We achieve higher performance even compared with some supervised methods, like SemEXP and ZSON.
Note that the improvement of UniGoal on IIN is more significant than ON. This is because the goal in ON is a single object node, which cannot fully exploit the potential of UniGoal on graph matching and reasoning.
Moreover, UniGoal also achieves state-of-the-art performance among universal goal-oriented methods, even outperforming supervised methods like PSL and GOAT with $+3.9/1.0$ lead on ON, $+22.8/7.6$ lead on IIN and $+3.2/2.6$ lead on TN. 

It is worth noting that for ON, our goal graph degenerates to a single node and thus some modules in UniGoal will not work, such as the goal graph decomposition in stage 1 and anchor pair alignment in stage 2. In this case, UniGoal still outperforms the scene graph-based zero-shot method SG-Nav on all benchmarks, which validates the effectiveness of graph correction and goal verification method in stage 3.

\subsection{Ablation Study}
We conduct ablation experiments on HM3D to validate the effectiveness of each part in UniGoal. We report performance of the ablated versions on the representative IIN task. 

\begin{table}[t]
    \centering
    \caption{Effect of pipeline design in UniGoal on HM3D (IIN) benchmark.}
    \begin{tabular}{lcc}
        \toprule
        Method & SR & SPL \\
        \midrule
        Simplify graph matching & 54.9 & 20.7 \\
        Remove blacklist mechanism & 50.6 & 17.3 \\
        Simplify multi-stage exploration policy & 59.0 & 23.2 \\
        \textbf{Full Approach} & \textbf{60.2} & \textbf{23.7} \\
        \bottomrule
    \end{tabular}
    \label{tab:ablation_1}
\end{table}

\textbf{Pipeline Design.} In Table \ref{tab:ablation_1}, we ablate the graph matching method and multi-stage explore policy. 
We first simplify graph matching by removing the score computation. In this way, once anchor pair is matched, the agent will enter stage 2. Similarly, the agent will enter stage 3 once the central object $o$ of $\mathcal{G}_g$ is matched. It is shown that the agent cannot switch the exploration strategy at the optimal time without a judgment of matching degree, leading to more failed cases. Then we remove the blacklist mechanism and observe a significant performance degradation. This is due to repeated matching on some failed nodes and edges, which makes the agent get stuck.
Finally we simplify the multi-stage exploration policy by removing stage 2. The agent will directly switch from unknown region exploration to goal approaching according to the matching result. It is shown that the performance drops significantly when applying a simpler exploration policy.

\textbf{Effect of Each Exploration Stage.} As shown in Table \ref{tab:ablation_2}, we conduct ablation studies on each exploration stage. 
For stage 1, we first replace the whole stage with a simple FBE strategy. Then we remove the goal graph decomposition and prompt the LLM with whole graph. We also remove frontier selection by making LLM predict a point location of the goal rather than scoring each frontier. The results in the first three rows show that each component in stage 1 is effective.
For stage 2, we first simplify the LLM-based coordinate projection method to a random guess of the 2D coordinate. We then remove the anchor pair alignment method by directly making LLM predict the goal location based on the BEV graphs. The experimental results in the fourth and fifth rows validate the effectiveness of inferring the goal location with structure overlap between graphs.
For stage 3, we remove the scene graph correction and goal verification in turn and report the results in the sixth and seventh rows. It is shown that both method works well to improve our final performance, which means correcting the scene graph as well as verifying the goal graph are essential for robust navigation.

\begin{table}[t]
    \centering
    \caption{Effect of the submodules in each stage during multi-stage scene exploration on HM3D (IIN) benchmark.}
    \begin{tabular}{lccc}
        \toprule
        Method & Stage & SR & SPL \\
        \midrule
        Replace stage 1 with FBE & 1 & 55.1 & 20.8 \\
        Remove $\mathcal{G}_{g}$ decomposition & 1 & 59.2 & 22.6 \\
        Remove frontier selection & 1 & 57.4 & 22.0 \\
        Simplify coordinate projection & 2 & 59.1 & 22.7 \\
        Remove anchor pair alignment &2 & 58.9&  22.6\\
        Remove $\mathcal{G}_{t}$ correction & 3 & 59.5 & 23.5 \\
        Remove goal verification & 3 & 58.2 & 22.4 \\
        \textbf{Full Approach} & -- & \textbf{60.2} & \textbf{23.7} \\
        \bottomrule
    \end{tabular}
    \label{tab:ablation_2}
\end{table}

\begin{figure*}[t]
    \centering
    \includegraphics[width=\linewidth]{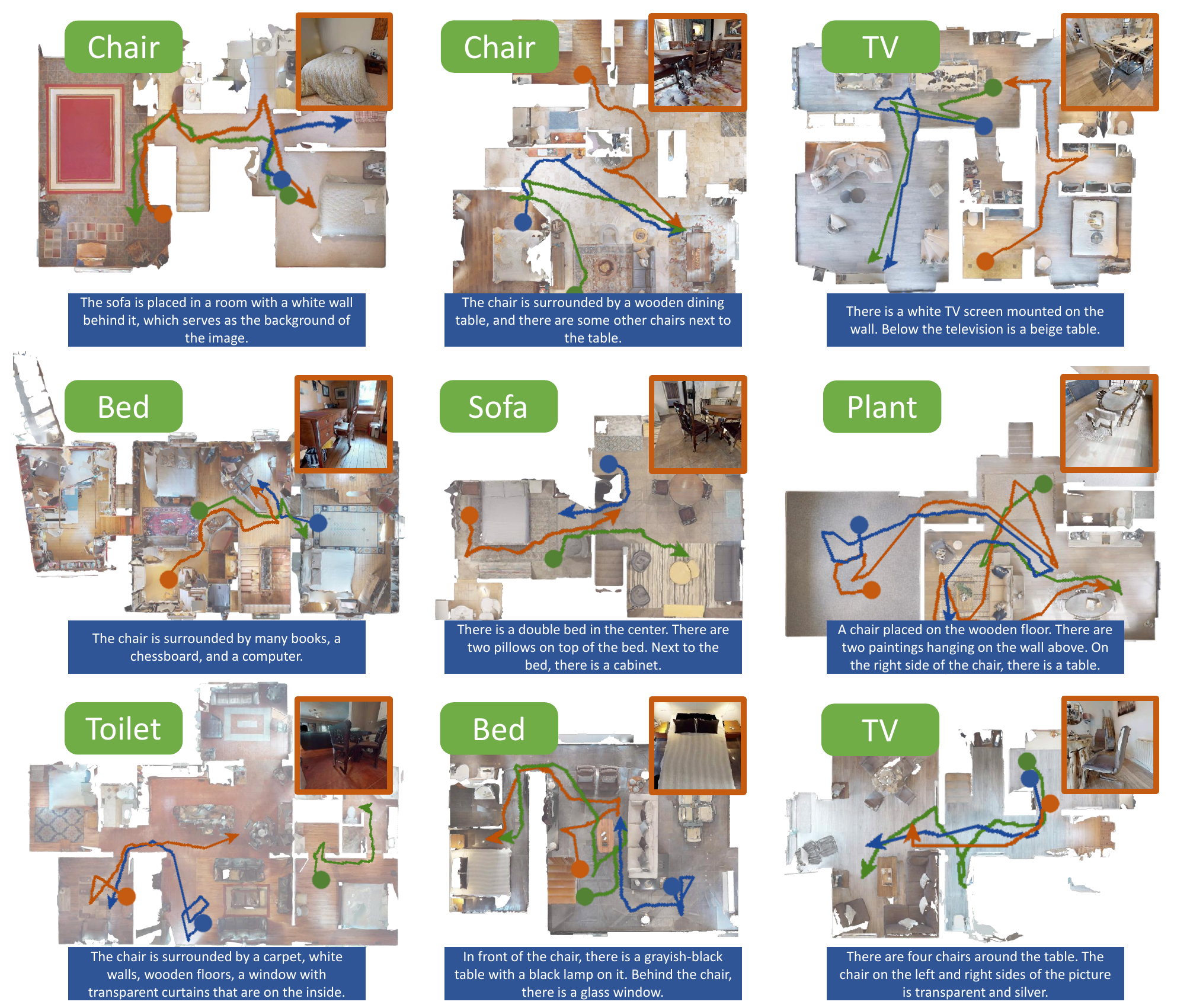}
    \caption{Visualization of the navigation path. We visualize ON (Green), IIN (Orange) and TN (Blue) path for several scenes. UniGoal successfully navigates to the target given different types of goal and diverse environments.}
    \label{fig:visualization_result}
\end{figure*}

\subsection{Qualitative Results}
We provide some qualitative results of UniGoal for better understanding. 
We first demonstrates the decision process of UniGoal in Figure \ref{fig:visualization_approach}. It is shown that UniGoal gradually increases the matching score by graph-based exploration. 
We also visualize the navigation path of UniGoal on the three studied tasks on 9 scenes of HM3D in Figure \ref{fig:visualization_result}. UniGoal can effectively handle all tasks with a single model, and generates efficient trajectory to reach the goal.


\section{Conclusion}
    In this paper, we have presented UniGoal, a universal zero-shot goal-oriented navigation framework which can handle object-goal navigation, instance-image-goal navigation and text-goal navigation in a single model without training or finetuning.
Since different types of goal usually requires totally different goal representation and LLM inference pipeline for zero-shot navigation, it is challenging to unify them with a single framework. To solve this problem, we convert the agent's observation into a scene graph and propose a uniform graph-based goal representation. In this way, the scene and goal are represented consistently, based on which we design a graph matching and matching-guided multi-stage scene exploration policy to make LLM fully exploit the correlation between scene and goal. Under different overlap, we propose different strategies to locate the goal position in the scene graph.
Besides, we maintain a blacklist that records unsuccessful matching to avoid repeated exploration.
Experimental results on three widely used datasets validate the effectiveness of UniGoal.
We further deploy UniGoal on real-world robotic platform to demonstrate its strong generalization ability and application value.


{
    \small
    \bibliographystyle{ieeenat_fullname}
    \bibliography{main}
}

\appendix
\clearpage
\setcounter{page}{1}
\maketitlesupplementary

\section{Overview}
\label{sec:overview}
This supplementary material is organized as follows:
\begin{itemize}
\item Section \ref{sec:subtask} provides the details of the three studied tasks.
\item Section \ref{sec:alg} provides the overall pipeline of UniGoal in algorithm for better understanding.
\item Section \ref{sec:detail} provides details on the approach.
\item Section \ref{sec:prompt} details the prompts for LLM.
\end{itemize}

\section{Definition of Each Task}
\label{sec:subtask}
We provide samples of the goal in each task in Table \ref{tab:sample} for better understanding. 
The goal for ON is a category in text format. The goal for IIN is an image with $o$ located at the center. For TN, the goal is a description about $o$, such as its relationship with other relevant objects in the scene.





\begin{table}[]
    \centering
    \begin{tabular}{c|M{0.22\linewidth}M{0.22\linewidth}M{0.22\linewidth}}
        \toprule
        ON & \textbf{Plant} & \textbf{Chair} & \textbf{Toilet} \\
        \midrule
        \multirow{2}*{IIN} & \includegraphics[width=\linewidth]{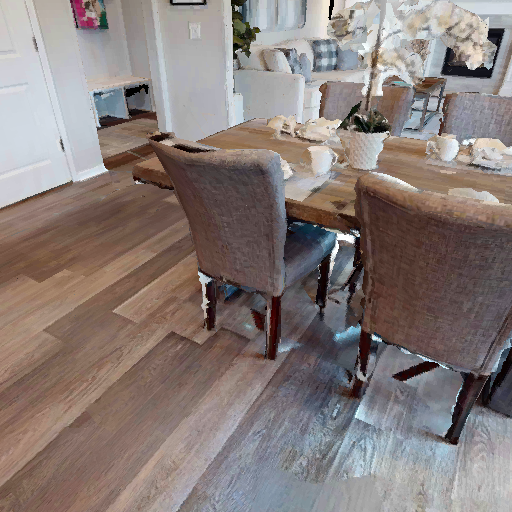} & \includegraphics[width=\linewidth]{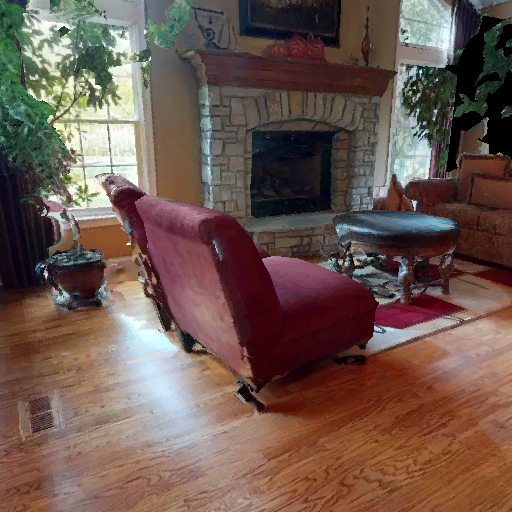} & \includegraphics[width=\linewidth]{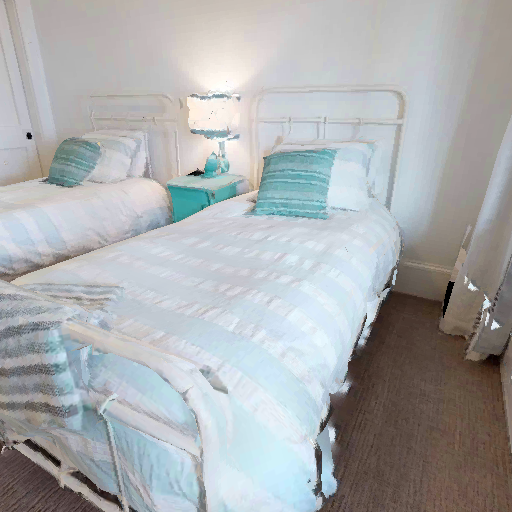} \\
         & \textbf{Chair} & \textbf{Sofa} & \textbf{Bed} \\
        \midrule
        TN & The \textbf{\textcolor{red}{toilet}} in this image is white, surrounded by a white door, beige tiles on the walls and floor. & The \textbf{\textcolor{red}{bed}} has white bedsheets. The bedroom has a double bed, two pillows and blankets, a chair and a table. & The \textbf{\textcolor{red}{chair}} is yellow and covered with red floral patterns. There is a wooden dining table in the upper left corner. \\
        \bottomrule
    \end{tabular}
    \caption{Illustration of goal in each task, with central objects colored in red.}
    \label{tab:sample}
\end{table}

\section{Pipeline of UniGoal}
\label{sec:alg}

\begin{algorithm}[h]
\caption{Overall Pipeline of UniGoal}
\begin{algorithmic}
\REQUIRE{Goal $g$, Observation $\mathcal{I}$}
\ENSURE{Goal Position $(x, y)$}
\STATE $\mathcal{G}_{t}\leftarrow \textrm{NewSceneGraph}()$
\STATE $\mathcal{G}_{g}\leftarrow \textrm{ConstructGoalGraph}(g)$
\STATE $\mathcal{B} \leftarrow$ NewBlackList()
\WHILE{True}
        
    \STATE $\mathcal{G}_{t} \leftarrow$ UpdateSceneGraph($\mathcal{G}_{t},\mathcal{I}$)
    \STATE $S,\mathcal{M}_N,\mathcal{M}_{E} \leftarrow$ GraphMatching($\mathcal{G}_{t}$, $\mathcal{G}_{g}$)
    \STATE Stage2 $\leftarrow$ $(\sigma_{1}<=S<\sigma_{2}$ and $\left |\mathcal{M}_{N}\right |>=2)$
    \STATE Stage3 $\leftarrow$ $(S>\sigma_{2}$ and $o \in \mathcal{M}_{N})$
    \STATE $i \leftarrow 0$
    \vspace{-2mm}
    \begin{graybox}
        \textcolor{gray}{============== \texttt{Stage 1} ==============}
        \vspace{-4.5mm}
        \IF{not (Stage2 or Stage3)}
            \STATE $\mathcal{G}_{g}^{sub} \leftarrow$ GraphDecomposition($\mathcal{G}_{g}$)
            \STATE $(x,y) \leftarrow$ SearchFrontier($\mathcal{G}^{sub}_{g}$,~$\mathcal{G}_{t}\setminus\mathcal{B}$)
            \STATE Go to $(x,y)$
        \ENDIF
    \end{graybox}
    \vspace{-4.5mm}
    \begin{graybox}
        \textcolor{gray}{============== \texttt{Stage 2} ==============}
        \vspace{-4.5mm}
        \IF{Stage2}
            \STATE $\mathcal{G}_{t}',~\mathcal{G}_{g}' \leftarrow$ CoordinateProjection($\mathcal{G}_{t} \setminus \mathcal{B},~\mathcal{G}_{g}$)
            \STATE $\mathbb{P}\leftarrow$ AnchorPairAlign($\mathcal{M}_N,\mathcal{M}_{E},~\mathcal{G}_{t}'$,~$\mathcal{G}_{g}'$)
            \STATE \textcolor{gray}{\texttt{\#~$\mathbb{P}=\{(x_i,y_i)|i=1,2,\dots,n\}$}}
            \IF{$1<=i<=n$}
                \STATE Go to $(x_i,y_i)$
                \STATE $i \leftarrow i+1$
            \ELSE
                \STATE $\mathcal{B} \leftarrow$ UpdateBlackList($\mathcal{M}_N,\mathcal{M}_{E},\mathcal{B}$)
                \STATE $i\leftarrow 0$
            \ENDIF
        \ENDIF
    \end{graybox}
    \vspace{-4.5mm}
    \begin{graybox}
        \textcolor{gray}{============== \texttt{Stage 3} ==============}
        \vspace{-4.5mm}
        \IF{Stage3}
            \STATE $\mathcal{G}_{t} \leftarrow$SceneGraphCorrection($\mathcal{G}_{t}$,~$\mathcal{I}$)
            \IF{GoalVerification($\mathcal{G}_{g},\mathcal{I}$)}
                \STATE Go to $(x,y)$  \textcolor{gray}{\texttt{~~\#~Navigation Stop}}
            \ELSE
                \STATE $\mathcal{B} \leftarrow$ UpdateBlackList($\mathcal{M}_N,\mathcal{M}_{E},\mathcal{B}$)
            \ENDIF
        \ENDIF
    \end{graybox}
    \vspace{-2.5mm}
\ENDWHILE
\end{algorithmic}\label{alg:block}
\end{algorithm}

We provide the algorithm diagram of UniGoal in Algorithm \ref{alg:block}. One gray box represents a stage.

\section{Details of Approach}
\label{sec:detail}

\subsection{Goal Graph Construction}

For Object-goal, we simply construct a graph with only one node and no edge, where the content of the node is the category of goal. For Instance-image-goal, we adopt Grounded-SAM~\cite{li2022grounded} to identify all the objects in the image, and then we prompt VLM~\cite{liu2023llava} to identify the relationships between objects to construct edges. For Text-goal, we prompt LLM to first identify all objects in the description, and then generate relationships between objects.

\subsection{Graph Embedding}
To embed nodes ($\mathcal{V}_{t}$ and $\mathcal{V}_{g}$), the embedding function is implemented as ${\rm Embed}(v)={\rm concat}({\rm CLIP}(v),{\rm Degree}(v))$, where ${\rm CLIP}(\cdot)$ is CLIP~\cite{radford2021learning} text encoder, ${\rm Degree}(\cdot)$ is the degree of a node.
To embed edges ($\mathcal{E}_{t}$ and $\mathcal{E}_{g}$), the embedding function is implemented as ${\rm Embed}(e)={\rm CLIP}(e)$.


\subsection{Zero Matching}
We detail the LLM-guided graph decomposition and frontier score computation used in zero-matching stage in this subsection. 

\noindent\textbf{LLM-guided Graph Decomposition.} We decompose the $\mathcal{G}_{g}$ into several subgraphs $\mathcal{G}_{g}^{i}$ by dividing $\mathcal{V}_{g}$ and assign the edges into each subgraph. The node sets $\mathcal{V}_{g}^{i}$ of all subgraphs comprise a division of $\mathcal{V}_{g}$, i.e., they satisfy:
\begin{align*}
\bigcup_{i} \mathcal{V}_{g}^{i}&=\mathcal{V}_{g} \\
\mathcal{V}_{g}^{i}\cap\mathcal{V}_{g}^{j}&=\emptyset~~~~\forall{i\neq j}
\end{align*}
We prompt LLM to conduct the division of $\mathcal{V}_{g}$. Then we assign all the edges whose connected two nodes are in the same $\mathcal{V}_{g}^{i}$ into the subgraph $\mathcal{G}_{g}^{i}$. The edges whose connected two nodes are not in a same $\mathcal{V}_{g}^{i}$ will be excluded during decomposition.

\noindent\textbf{Frontier Scoring.} Following SG-Nav, we prompt LLM with chain-of-thought (CoT) to predict the most likely position of $\mathcal{V}_{g}^{i}$ and score the frontiers according to their distance to the most likely position and their distance to the agent. Then we select the center of frontier with maximum score as the long-term goal.

\subsection{Hyperparameters}
The maximum navigation step number $T$ is set as: $T=500$ for ON, $T=1000$ for IIN and TN. The distance of success condition $r$ is set as: $r=1.6$m for ON, $r=1.0$m for IIN and TN. The threshold of similarity for matching pairs of nodes and edges $\tau$ is set as: $\tau=0.9$. The matching score thresholds are set as: $\sigma_1=0.5,\sigma_2=0.9$.


\section{Prompts}
\label{sec:prompt}

We provide all prompts used in UniGoal.

\noindent\textbf{Goal Construction for IIN:}

\begin{graybox}
    \texttt{You are an AI assistant that can infer relationships between objects. You need to guess the spatial relationship between \textcolor{red}{\{object1\}} and \textcolor{red}{\{object2\}} in the \textcolor{red}{\{image\}}. Answer relationship with one word or phrase.}
\end{graybox}

\vspace{-2mm}
where \{object1\}, \{object2\} and \{image\} will be replaced by the two objects and the goal image.

\noindent\textbf{Goal Construction for TN:}

\begin{graybox}
    \texttt{You are an AI assistant that can identify objects and relationships from a description. You need to list the objects and relationships in \textcolor{red}{\{text\}} in following format:}
    
    \texttt{\{'nodes': [\{'id': 'book'\}, \{'id': 'table'\}],}
    
    \texttt{'edges': [\{'source': 'book', 'target': 'table', 'type': 'on'\},]}
\end{graybox}

\vspace{-2mm}
where \{text\} will be replaced by the goal text.

\noindent\textbf{Goal Decomposition:}

\begin{graybox}
    \texttt{You are an AI assistant with commonsense. You need to divide the following objects \textcolor{red}{\{\}} into subsets based on correlation, with objects within each subset strongly correlated and objects within different subsets not strongly correlated.}
    
    \texttt{Your response should be in two-dimensional array format: \textcolor{red}{[[object1, object2,\dots,objectn], [\dots],\dots,[\dots]]}}
\end{graybox}

\vspace{-2mm}
where \{\} will be replaced by the objects in $\mathcal{V}_{g}$.

\noindent\textbf{Frontier Scoring:}

\begin{graybox}
    \texttt{You are an AI assistant with commonsense. You need to predict the most likely distance between the \textcolor{red}{\{object\}} and the \textcolor{red}{\{subgraph\}}. Answer a  distance number in meter.}
    
\end{graybox}

\vspace{-2mm}
where \{object\} and \{subgraph\} will be replaced by each object in $\mathcal{G}_{t}$ and the description of decomposed $\mathcal{G}_{g}$.


    

    
    
    
    


\noindent\textbf{Scene Graph Correction:}

\begin{graybox}
    \texttt{You are an AI assistant with commonsense and strong ability to give a more detailed description of a node or edge in an indoor scene graph.}
    
    \texttt{Now give a more detailed description of \textcolor{red}{\{\}} based on the graph \textcolor{red}{\{graph\}} and the newly observed image \textcolor{red}{\{image\}} in order to identify possible errors in the scene graph.}

    
    
    
    
\end{graybox}

\vspace{-2mm}
where \{\}, \{graph\} and \{image\} will be replaced by a node or edge, the local graph $\mathbf{A} \cdot \mathcal{V}^{(t)}_{o},\mathbf{M} \cdot \mathcal{E}_{o}^{(t)}$ or $\mathbf{M}^{T} \cdot \mathcal{V}^{(t)}_{o},\mathbf{A}' \cdot \mathcal{E}_{o}^{(t)}$ and the description of the newly observed image.




\end{document}